\documentclass[conference]{IEEEtran}
\IEEEoverridecommandlockouts
\usepackage[square,sort,comma,numbers]{natbib}
\usepackage{amsmath,amssymb,amsfonts}
\usepackage{algorithmic}
\usepackage{graphicx}
\usepackage{textcomp}
\usepackage{xcolor}
\usepackage{url}
\usepackage{booktabs}

\def\BibTeX{{\rm B\kern-.05em{\sc i\kern-.025em b}\kern-.08em
    T\kern-.1667em\lower.7ex\hbox{E}\kern-.125emX}}
\begin{document}

\title{Tri-VQA: Triangular Reasoning Medical Visual Question Answering for Multi-Attribute Analysis\\
}

\author{\IEEEauthorblockN{1\textsuperscript{st} Lin Fan}
\IEEEauthorblockA{\textit{School of Computing and Artificial Intelligence} \\
\textit{Southwest Jiaotong University}\\
Chengdu, China \\
linfan@my.swjtu.edu.cn}
\and
\IEEEauthorblockN{2\textsuperscript{nd} Xun Gong}
\IEEEauthorblockA{\textit{School of Computing and Artificial Intelligence} \\
\textit{Southwest Jiaotong University}\\
Chengdu, China \\
xgong@swjtu.edu.cn}
\and
\IEEEauthorblockN{3\textsuperscript{rd} Cenyang Zheng}
\IEEEauthorblockA{\textit{School of Computing and Artificial Intelligence} \\
\textit{Southwest Jiaotong University}\\
Chengdu, China \\
z\_c\_y@my.swjtu.edu.cn}
\and
\IEEEauthorblockN{4\textsuperscript{th} Yafei Ou}
\IEEEauthorblockA{\textit{Institute of Innovative Research} \\
\textit{Tokyo Institute of Technology}\\
Yokohama, Japan \\
ou.y.ac@m.titech.ac.jp}
}
\maketitle

\begin{abstract}
The intersection of medical Visual Question Answering (Med-VQA) is a challenging research topic with advantages including patient engagement and clinical expert involvement for second opinions. However, existing Med-VQA methods based on joint embedding fail to explain whether their provided results are based on correct reasoning or coincidental answers, which undermines the credibility of VQA answers. In this paper, we investigate the construction of a more cohesive and stable Med-VQA structure. Motivated by causal effect, we propose a novel Triangular Reasoning VQA (Tri-VQA) framework, which constructs reverse causal questions from the perspective of "Why this answer?" to elucidate the source of the answer and stimulate more reasonable forward reasoning processes. 
We evaluate our method on the Endoscopic Ultrasound (EUS) multi-attribute annotated dataset from five centers, and test it on medical VQA datasets. Experimental results demonstrate the superiority of our approach over existing methods. Our codes and pre-trained models are available at \url{https://anonymous.4open.science/r/Tri_VQA}.
\end{abstract}


\section{Introduction}
\begin{figure*}[!t]
    \centering
    \includegraphics[width=\textwidth]{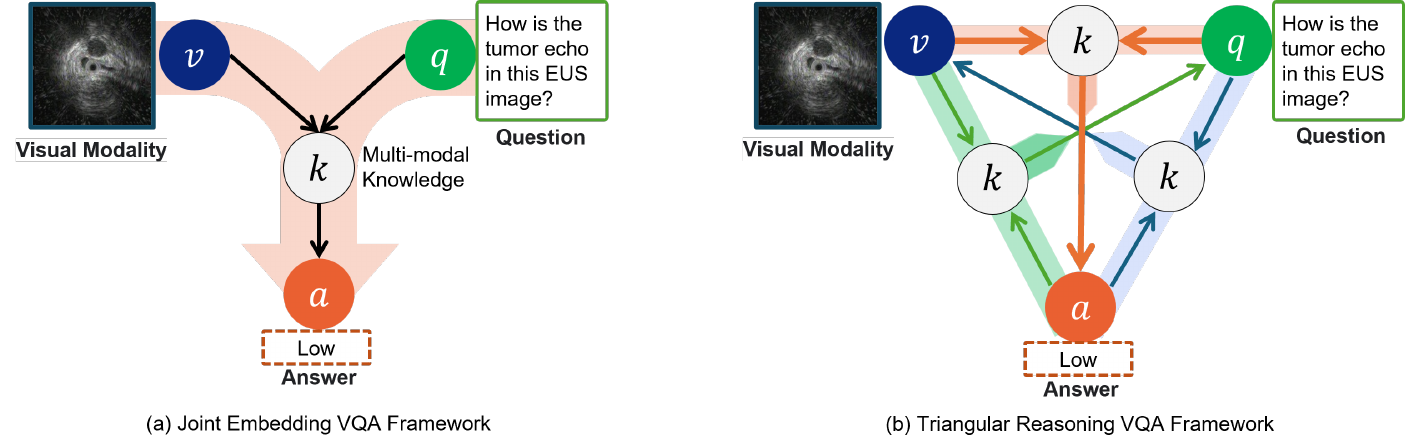}
    \caption{Joint Embedding VQA framework vs. Triangular Reasoning VQA framework. Tri-VQA utilizes mutual inference constraints among \textit{v} (visual), \textit{q} (question), and \textit{a} (answer) to explain the rationality of generated answers.} 
    \label{fig:1}
\end{figure*}

Visual Question Answering (VQA) is a multidisciplinary problem that combines Computer Vision (CV) and Natural Language Processing (NLP) to answer questions based on the content of an image. In recent years, there has been a growing interest in exploring VQA systems in the medical domain, known as Med-VQA, leveraging advancements made in general scenarios\cite{kovaleva2020towards}. Med-VQA systems have the potential to contribute to clinical decision-making and enhance patient engagement. Unlike other systems that focus on predefined diseases or organ types, medical VQA systems are capable of comprehending free-form questions in natural language and providing reliable and user-friendly answers.  

The predominant approach in Med-VQA is the employment of a joint embedding framework, which integrates extracted image features with question features to predict or generate an answer~\cite{kim2016multimodal,fukui2016multimodal}. Currently, researchers are primarily focused on investigating enhanced fusion algorithms or augmenting answer accuracy through the examination of questions in medical datasets, such as attention mechanisms~\cite{gong2021cross,liu2019effective} and early question analysis~\cite{liu2022medical,vu2020question}. However, the achieved performance still remains unsatisfactory. The diversity of medical imaging modalities and disease characteristics demands Med-VQA models to possess higher-level reasoning capabilities~\cite{jifara2019medical,xi2020visual}. To the best of our knowledge, there is currently no specialized Med-VQA method specifically trained for tumor attributes. The automatic recognition of tumor attributes is crucial as it provides detailed tumor information for accurate diagnoses and aids in model training by mimicking the clinical process and offering interpretability \cite{manh2022multi}. The limited reasoning capabilities of existing Med-VQA methods may be a significant factor hindering research on precise and fine-grained attribute analysis. Moreover, many studies have not focused on the deep relationship between questions and images, i.e., whether the results they provide are based on correct reasoning or coincidental answers~\cite{wang2020general}. This creates ambiguity regarding the reliability of Med-VQA.

In this paper, we aim to enable attribute recognition in medical images and train a fine-grained VQA system with multiple attributes. Existing joint embedding VQA methods focus on forward causal questions ${(Q+V \rightarrow A)}$, ignoring the "Why this answer??". Inspired by Gelman et al.\cite{NBERw19614}, we propose incorporating reverse causal questions to stimulate correct reasoning. For instance, given an answer, we construct reverse causal reasoning to speculate the image's contents or the question. This approach enhances the reasoning structure, enabling the network to better understand the interplay and achieve a more seamless multimodal fusion. Furthermore, the accuracy of reverse inference serves as a reliable indicator for the reasoning of answers. We introduce the Tri-VQA(shown in Fig. \ref{fig:1}), which incorporates medical multi-attribute information, providing robust reasoning and reliability in Med-VQA.  Our contributions are summarized as follows:
\begin{enumerate}
    \item We propose a novel triangular reasoning model, named Tri-VQA. Tri-VQA addresses the limitation of existing Med-VQA methods that solely consider forward causal reasoning by introducing reverse causal questions to promote accurate reasoning. This improvement enhances the stability of the reasoning structure and facilitates the formation of reliable answers based on reasoning.
    \item We provide a potential indicator for assessing the reliability of answers in Med-VQA through the analysis of the accuracy of reverse inference.
    \item Our Tri-VQA model achieves superior performance on medical VQA benchmarks with open-ended questions. Furthermore, experiments conducted on the EUS dataset validate the effectiveness of our proposed model in multi-attribute analysis.
\end{enumerate}
\begin{figure*}[!t]
    \centering
    \includegraphics[width=\textwidth]{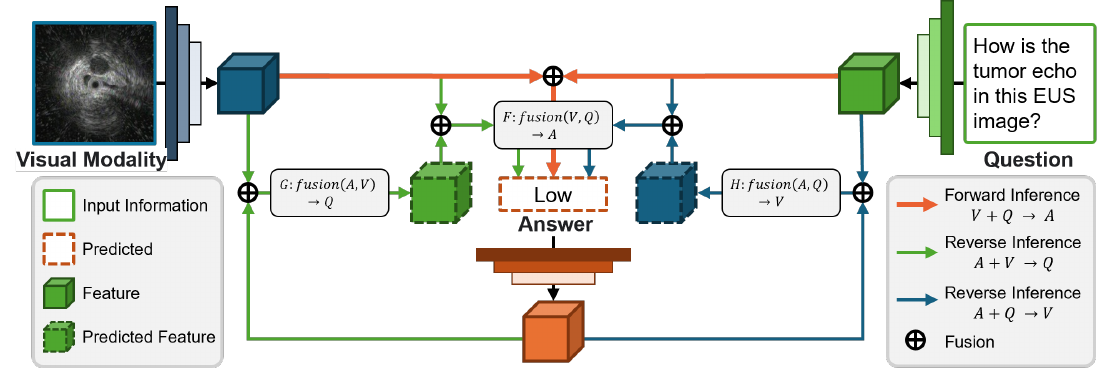}
    \caption{The overall framework of Tri-VQA. We perform forward inference using two input information sources (represented by orange arrows) to obtain the inference for the answer. Subsequently, we utilize the predicted answer to perform backward inference for image features or question features (represented by green and blue arrows, respectively). The predicted features are constrained by a similarity constraint with the ground truth features. Both sets of features generated through backward inference are then fed into the forward inference function ${F}$ to infer the final answer, which is further constrained by the true label.}
    \label{fig:all}
\end{figure*}

\section{Related Word}
\subsection{Medical Visual Question Answering}
In recent years, the fusion of CV and NLP in fields like Med-VQA has gained attention. The joint embedding framework, including image and question feature extractors (e.g., VGG and ResNet for images, LSTM and BERT for questions), fusion networks, and answer generation networks, is commonly used in Med-VQA\cite{kim2016multimodal,fukui2016multimodal}. Fusion strategies like QC-MLB\cite{vu2020question} and MedFuseNet\cite{sharma2021medfusenet}, based on attention mechanisms, have been proposed for enhancing the interaction between visual and textual modalities. However, variations in medical imaging modalities (CT, MRI, X-ray, ultrasound) and differences in indicator presentation across medical conditions necessitate higher-level reasoning capabilities in Med-VQA models\cite{jifara2019medical,xi2020visual}. Despite efforts to address these challenges and improve performance\cite{nguyen2019overcoming,zhan2020medical}, performance levels remain low. Moreover, there is a lack of evidence to demonstrate that the answers provided are based on a comprehensive analysis of both the question and visual information.
\section{Methods}
The recognition of tumor attributes plays a crucial role in assisting doctors in making accurate diagnoses~\cite{manh2022multi,zheng2023enhancing}. Additionally, it provides procedural diagnostic information, thereby enhancing the credibility of the model. The flexibility and adaptability of Visual Question Answering (VQA) systems not only improve interactions between doctors and the system, leading to increased work efficiency, but also enable self-service for patients, providing them with additional support. Following the standard formulation, we define the VQA task as a multi-class classification problem. Figure~\ref{fig:all} illustrates the VQA framework we have developed for the multi-attribute analysis of EUS tumors. Given an EUS image $V = v$ with answers $A = a_1,...,a_6$ and corresponding attribute questions $Q = q_1,...,q_6$, we first establish a conventional pathway (\S\ref{sec:vqa}) to generate $a_{pre}$ by reasoning over $v$ and $q$. Subsequently, we propose the Tri-VQA framework (\S\ref{sec:avq}), which involves reverse reasoning of $q_{pre}$ and $v_{pre}$ based on $a$, enabling a comprehensive understanding among the three components and constraining the generation of the most plausible $a_{pre}$. Specifically, Tri-VQA involves three types of reasoning: forward reasoning $F: fusion(Q, V) \rightarrow A$, and two types of reverse reasoning $G: fusion(A, V) \rightarrow Q$ and $H: fusion(A, Q) \rightarrow V$. Tri-VQA allows for mutual reasoning among these three components to constrain the validity of the answers and provide justifications.
\subsection{Forward Inference: ${V+Q\rightarrow A}$}
\label{sec:vqa}
This module consists of three sequentially organized components: multimodal feature extraction module, multimodal feature fusion module, and answer prediction module. We transform VQA into a classification problem based on different attribute questions, where each classification task consists of ${C}$ candidate answers, and ${C}$ can be different. Denote by ${D={(v_i,q_i,a_i)}^{n}_{i=1}}$ the training dataset for a VQA model, where ${n}$ is the number of training examples, and ${v}$, ${q}$, and ${a}$ denote the image, question and answer of a task respectively. The process of generating various attribute features in the images is outlined as follows:
\begin{equation}
    f^{PD}_{i}=fusion(F_{clip}(q_i),F_{bert}(v_i))
\end{equation}
where $i \in 1, ..., 6$ represents the i-th diagnostic attribute of EUS images, $f_i^{PD}$ denotes the multi-modal fusion feature corresponding to the ${i-th}$ attribute, $q_i$ represents the textual question posed for the ${i-th}$ attribute, $F_{clip}(\cdot)$ and $F_{bert}(\cdot)$ respectively denote the feature extractors for the questions and  images, and $fusion(\cdot)$ represents the feature fusion function, the choice of the function can be additive, concatenation, or any other multi-modal feature fusion method. Thus, we completed the forward reasoning for each attribute: ${F: f^{PD}{i} \rightarrow a_{pre, i}}$. Then, we combined ${f^{PD}{i}}$ for the final pathological diagnosis.

\subsection{Reverse Inference: ${A+V\rightarrow Q}$ and ${A+Q\rightarrow V}$}
\label{sec:avq}
We believe that the approach of reverse reasoning is crucial in the learning process of the Med-VQA. From the perspective of the model, the success of reverse reasoning represents a thorough understanding of the fusion features of visual information (\textit{v}), textual information (\textit{q}), and answers (\textit{a}). It helps to avoid the generation of unreasonable answers caused by a single path in the joint embedding model. From the perspective of the users, reverse reasoning provides an explanation for answer generation, demonstrating that the model has carefully considered the generated answer rather than relying on a random process.

\textbf{Specific Implementation of ${A+V\rightarrow Q}$}. ${A+V\rightarrow Q}$ is achieved by utilizing ${G}$ to infer potential question features based on the generated answer and visual pair ${(v_i,a_{pre,i})}$. Then, the inferred question features are constrained to be close to the actual question features. Instead of reverse mapping the inferred text features back to text, we directly compare the inferred question features to the ground truth question features, simplifying the implementation and reducing computational complexity. The loss definition for reverse reasoning of ${Q}$ is as follows:
\begin{equation}
L_{av\rightarrow q} = \sum_{i=1}^{n} (G(fusion(a_{pre,i}, v_i)) - q_i)^2
\end{equation}
Considering the potential semantic loss caused by enforced similarity constraints in inferring questions, which could render the reverse reasoning process meaningless, we introduce a secondary forward reasoning (SFR) to mitigate this issue and reinforce the generation of accurate answers. The SFR helps to enhance the semantic correctness of reverse reasoning for Q using the generated question and correct visual features. Additionally, we impose the constraint that the results of the SFR should match those of the initial forward reasoning (IFR), thus ensuring answer correctness. The loss definition for answer correctness involved in this part is as follows:
\begin{equation}
F(G(fusion(a_{pre,i}, v_i)),v_i) \overset{\text{CE}}\Longleftrightarrow a_{pre,i} \overset{\text{CE}}\Longleftrightarrow a_i
\end{equation}
where CE represents the use of cross-entropy loss,  ${a_{pre,i}}$ is generated by ${F(fusion(q_i, v_i))}$, ${a_i}$ represents the true label of the answer.

\textbf{Specific Implementation of ${A+Q\rightarrow V}$}. Similar to ${A+V\rightarrow Q}$, ${A+Q\rightarrow V}$ is achieved by utilizing ${H}$ to infer visual key features based on the generated answer and visual input ${(q_i,a_{pre,i})}$, and then comparing the inferred visual features with the ground truth visual features. The loss definition for reverse reasoning of ${V}$ is as follows:
\begin{equation}
L_{aq\rightarrow v} = \sum_{i=1}^{n} (H(fusion(a_{pre,i}, q_i)) - v_i)^2
\end{equation}
Symmetrically to the reverse reasoning for Q, we also impose semantic control on the generated ${v_{i,pre}}$:
\begin{equation}
F(H(fusion(a_{pre,i}, q_i)),q_i) \overset{\text{CE}}\Longleftrightarrow a_{pre,i} \overset{\text{CE}}\Longleftrightarrow a_i
\end{equation}
It is worth noting that our proposed semantic modulation process for reverse reasoning does not require any additional annotations and also enhances the training of the forward reasoning process of ${F: fusion(Q, V) \rightarrow A}$.

\section{Experiments}
\subsection{Setup}

\begin{table}[!t]
\caption{The number of questions and the number of unique answers for each type of question in the SLAKE dataset and the training and test sets.}
\centering
\resizebox{\linewidth}{!}{
\begin{tabular}{ccccc}
\toprule
& \multicolumn{2}{c}{Train Set} & \multicolumn{2}{c}{Test Set} \\ \cmidrule(lr){2-3} \cmidrule(lr){4-5}
Question Categories & Question  & Answer & Question & Answer \\ \midrule
what mod      & 2         & 3      & 2        & 3 \\
which organ      & 2         & 4      & 1        & 3 \\
which part    & 5         & 8      & 4        & 7      \\ how much         & 1         & 3      & 1        & 3      \\
what is       & 82        & 73     & 38       & 42     \\ what kind        & 14        & 10     & 6        & 6      \\
where is      & 58        & 48     & 30       & 27     \\ what disease     & 14        & 12     & 6        & 8      \\
what diseases & 1         & 30     & 1        & 16     \\ what imaging     & 1         & 3      & 1        & 2      \\
how many      & 19        & 7      & 13       & 7      \\ where does       & 2         & 5      & 2        & 4      \\
what organ    & 27        & 20     & 7        & 8      \\ how was          & 1         & 3      & 1        & 3      \\
what are      & 4         & 2      & 2        & 1      \\ in what          & 1         & 2      & 1        & 2      \\
where are     & 3         & 6      & 1        & 3      \\ what scanning    & 1         & 2      & 1        & 2      \\
which is      & 5         & 8      & 3        & 3      \\ what type        & 2         & 5      & 2        & 3      \\
what does     & 2         & 2      & 0        & 0      \\ which side       & 2         & 4      & 1        & 2      \\
which lobe    & 4         & 3      & 1        & 1      \\ what role        & 1         & 2      & 1        & 1      \\
what part     & 9         & 7      & 3        & 3      \\ what system      & 1         & 1      & 0        & 0      \\
which organs  & 14        & 29     & 14       & 14     \\ what density     & 1         & 1      & 1        & 1      \\
what color    & 34        & 5      & 12       & 3      \\ which hemisphere & 1         & 3      & 0        & 0      \\
how to        & 19        & 14     & 12       & 9      \\ which place      & 2         & 3      & 0        & 0      \\
what can      & 1         & 1      & 1        & 1      \\ what tissue      & 1         & 1      & 0        & 0     \\ \bottomrule
\end{tabular} %
}
\label{tab:slake}
\end{table}

\begin{table}[!t]
    \centering
    \caption{The number of questions and the number of unique answers for each type of question in the med-VQA dataset and the training and test sets.}
\resizebox{\linewidth}{!}{
\begin{tabular}{ccccc}
\toprule
&\multicolumn{2}{c}{Train Set} &\multicolumn{2}{c}{Test Set} \\ \cmidrule(lr){2-3} \cmidrule(lr){4-5}
Question Categories &Answer &Question &Answer &Question \\ \midrule 
MODALITY&47 &651 &47 &279   \\ 
ORGAN&25 &294 &25 &126  \\ 
POS&137 &1890 &137 &810 \\ 
OTHER&109 &924 &109 &396  \\ 
PLANE&10 &427 &10 &183  \\ 
ABN&46 &546 &46 &234 \\ 
PRES&110 &1274 &110 &546 \\ 
ATTRIB&27 &266 &27 &114 \\ 
PRES, POS&3 &28 &3 &12 \\ 
COLOR&9 &84 &9 &36 \\ 
COUNT&5 &84 &5 &36 \\ 
SIZE&11 &105 &11 &45 \\ 
POS, PRES&2 &28 &2 &12 \\ 
POS, ABN&1 &14 &1 &6  \\ 
POS, ABN&1 &7 &1 &3  \\ 
SIZE, COLOR&1 &14 &1 &6 \\ 
ATRIB&1 &7 &1 &3 \\ 
\bottomrule
\end{tabular} %
}
    \label{tab:rad}
\end{table}

\subsubsection{Datasets}
We conducted a comprehensive study on the Tri-VQA, using a multicenter EUS dataset. The dataset consists of 519 cases with a total of 7,021 images. Six attribute labels were extracted from diagnostic reports for each image. For the division of training and testing sets, we randomly selected 30\% of the data from each center according to the proportion of GISTs and non-GISTs as the testing set, and the remaining data as the training set. The EUS dataset used in this study includes EUS images acquired from six different EUS machines, including Olympus UM-2R 12MHZ, UM-2R 20MHZ, UM-3R 12MHZ, UM-3R 20MHZ, IM-02P 12MHZ, and IM-02P 20MHZ. The data were collected from real clinical environments. To thoroughly investigate the effectiveness of Tri-VQA, we conducted extensive experiments on two publicly available medical VQA datasets, VQA-RAD~\cite{lau2018dataset} and SLAKE~\cite{liu2021slake}. For both datasets, we chose only the more challenging "open-ended" questions for our experiments, as the answers to these questions are more semantically meaningful. We divided the training and test sets of the VQA-RAD and SLAKE datasets, Please refer to Table. ~\ref{tab:slake} and Table. ~\ref{tab:rad} for details regarding the setup of these datasets.

\subsubsection{Implementation details }
The image storage format of EUS is bmp format, and all images all read according to RGB format. Images were scaled down to 224 pixels on the short side and center cropped using a square of size 224 × 224 to maintain their original aspect ratio. Each pixel point in the image was normalized. We utilized a pre-trained CLIP (ViT-B/32)~\cite{radford2021learning} as the image feature extraction model. The BERT model used consists of 12 encoding layers, with each layer having 12 attention heads and a hidden layer of 768 dimensions. The initial learning rate for each layer was set to 0.001, with a momentum of 0.9, and the learning rate was reduced by a factor of 10 after 10 epochs. The network was optimized using stochastic gradient descent (SGD). Model training and evaluation were implemented in the PyTorch framework using the Geforce RTX 4090 GPU.
\begin{table}[!t]
    \centering
\caption{Performance comparison with state-of-the-art EUS diagnostic methods. "Multi-attr" represents whether to use multi-attribute analysis.}
\resizebox{\linewidth}{!}{
    \begin{tabular}{cccccc}
    \toprule
        Method & SEN &SPE &ACC &AUC  &Multi-attr  \\ \midrule
        Voice-Assisted \cite{bonmati2021voice} &0.740 &- & 0.7600 &-  & No   \\ 
        CNN-based \cite{seven2022differentiating} &0.920 &0.643 &0.869 &- & No \\ 
	MMP-AI \cite{zhu2022multimodal} &0.836 &0.833 & 0.835 &0.896 & No \\ 
	MAA-Net \cite{manh2022multi} & 0.868    &0.921     & 0.891    & 0.865    & Yes    \\ 
	Query2 \cite{he2023query2}  &0.888 &0.865 & 0.877 &-  & No \\ 
        Tri-VQA (Ours) & \textbf{0.962} & \textbf{0.944} & \textbf{0.957} & \textbf{0.935} & \textbf{Yes} \\ \bottomrule
    \end{tabular}
}
\label{table:eus}
\end{table}

\begin{table}[!t]
\caption{Comparison of Performance with State-of-the-Art Med-VQA Methods on EUS Multi-attribute Dataset.}
\resizebox{\linewidth}{!}{
\begin{tabular}{ccccccc}
\toprule
Method   & Echo & Boundary & Shape & Original & Extrude & Het- \\ \midrule
SAN \cite{lau2018dataset,yang2016stacked} & 0.799     &0.781   &0.754  &0.712  & 0.735  &0.768               \\
MCB \cite{lau2018dataset,fukui2016multimodal}     &0.801 &0.790 &0.743 &0.728  & 0.767 &0.783   \\
BAN \cite{kim2018bilinear,nguyen2019overcoming}     &0.823 &0.812&0.779&0.758&0.764& 0.794    \\
MEVF+SAN \cite{nguyen2019overcoming} &0.831 &0.820&0.754&0.789 &0.768&0.802  \\
MEVF+BAN \cite{nguyen2019overcoming} &0.837& 0.822 & 0.779& 0.768 &0.794 & 0.805    \\
Med-VQA \cite{liu2022medical}  &0.868 &0.844 &0.821 & 0.785 &0.801 & 0.837   \\
MUMC \cite{li2023masked} &0.911 &0.899& 0.842&0.869&0.835&0.882 \\ 
Tri-VQA (Ours)   &\textbf{0.981} &\textbf{0.991} &\textbf{0.924}&\textbf{0.921} &\textbf{0.926} &\textbf{0.941}   \\ \bottomrule
\end{tabular}}
\label{tab:san}
\end{table}

\subsection{Evaluation of Tri-VQA on a Multi-Attribute EUS Dataset}
To validate the benefits of Tri-VQA multi-attribute analysis for EUS pathology diagnosis, we compared it with the latest methods that utilize EUS for gastrointestinal stromal tumor (GIST) diagnosis. The results are presented in Table~\ref{table:eus}, where the MAA-Net results are obtained from training on the EUS dataset. The results clearly indicate that the utilization of multi-attribute analysis methods outperforms any other non-attribute analysis methods, even when employing multimodal approaches such as Voice-Assisted, which combines voice and image modalities for diagnosis. Additionally, the results of Tri-VQA surpass those of MAA-Net, which also incorporates multi-attribute analysis. Our analysis suggests that the superiority of Tri-VQA over MAA-Net can be attributed to the adoption of a triangular reasoning question-and-answer format, enabling accurate inference for each attribute. Consequently, more precise attribute features are obtained, leading to improved diagnostic outcomes.

We further extended our efforts in the research by training EUS on several recent Med-VQA methods to enable multi-attribute recognition. A comprehensive comparison was then conducted with Tri-VQA. The obtained results, showcasing the recognition accuracy of different attributes, are summarized in Table~\ref{tab:san}. Specifically, lau et. al \cite{lau2018dataset} directly employed existing VQA models in the general domain to solve Med-VQA, e.g., the stacked attention networks (SAN) \cite{yang2016stacked} and the multimodal compact bilinear pooling (MCB) \cite{fukui2016multimodal}. To develop a specialized Med-VQA system, MEVF\cite{nguyen2019overcoming} proposed a mixture of enhanced visual features (MEVF) framework and combined it with different attention mechanisms such as bilinear attention networks (BAN) \cite{kim2018bilinear} and SAN. Med-VQA \cite{liu2022medical} utilizes conditional reasoning and contrastive learning to address Med-VQA problems by jointly modeling the question and image features, aiming to achieve better fusion between them. MUMC \cite{li2023masked} enhances the model's understanding of the correlation between images and language by employing masked pre-training. These methods aim to enhance the reasoning ability for answering questions by improving the correlation between question and image features. In contrast, the proposed Tri-VQA approach enhances the connections among Vision, Question, and Answer from the perspective of reverse reasoning. It leverages reverse reasoning constraints to promote the correctness and rationality of answers, thereby achieving the best results in more challenging fine-grained multi-attribute analysis tasks.

\begin{table}[!t]
\centering
\caption{Performance comparison with state-of-the-art Med-VQA methods}
\resizebox{\linewidth}{!}{
\begin{tabular}{cccccc}
\toprule
Open-ended&Text  &Vision  &Fusion&VQA-RAD&SLAKE\\ \midrule
\cite{zheng2020learning}   &Glove+GRU&GRU&BLOCK   &0.600         & -      \\
\cite{liu2022medical}   &LSTM+MLP&ResNet-50&CNN &0.605         & 0.805       \\
\cite{dhanush2021vqa} &BiLSTM&ResNet&CNN&0.678   &0.811       \\
\cite{wang2022m2fnet} &LSTM&Ensemble&PCBI&0.688   &-       \\
\cite{bazi2023vision} &ViT&Transformer&CONCAT& 0.729        & -      \\ 
\cite{huang2023medical} &BERT+GRU&ResNet-50&CONCAT&-        &0.806      \\  
Tri-VQA (Ours)   &BERT&CLIP & ADD & \textbf{0.738}   &\textbf{0.831}      \\ \bottomrule
\end{tabular}}
    \label{tab:med}
\end{table}

\begin{table*}[!t]
\caption{The measurement of inference reliability during testing for different attributes.}
\resizebox{\linewidth}{!}{
\begin{tabular}{ccccccccccccc}
\toprule
& \multicolumn{2}{c}{Echo} & \multicolumn{2}{c}{Boundary} &\multicolumn{2}{c}{Shape} &\multicolumn{2}{c}{Original} & \multicolumn{2}{c}{Extrude} & \multicolumn{2}{c}{Het-} \\ \cmidrule(lr){2-3} \cmidrule(lr){4-5} \cmidrule(lr){6-7} \cmidrule(lr){8-9} \cmidrule(lr){10-11} \cmidrule(lr){12-13}
&MSE &Euclidean &MSE &Euclidean &MSE &Euclidean &MSE &Euclidean &MSE &Euclidean &MSE &Euclidean \\ \midrule
\multicolumn{13}{c}{${A+V\rightarrow Q}$} \\ \midrule
Correct  &4.48&0.161 &4.31&0.232 &5.01&0.227 &4.52&0.265 &5.40&0.238 &4.62&0.180     \\ 
Incorrect  &5.26&1.970 &4.90&0.650 &7.24&0.803 &6.01&0.405 &6.52&0.558 &4.71&0.971   \\ \midrule
\multicolumn{13}{c}{${A+Q\rightarrow V}$} \\ \midrule
Correct  &2.35&0.009 &2.84&0.030 &4.72&0.087 &4.21&0.053 &3.57&0.076 &3.09&0.043    \\ 
Incorrect  &3.42&0.221 &4.23&0.138 &5.64&0.184 &4.17&0.064 &3.91&0.124 &5.23&0.138   \\ \bottomrule
\end{tabular}}
\label{tab:dis}
\end{table*}

\subsection{Evaluation of Tri-VQA on Public Datasets}
To comprehensively evaluate the advantages of the Tri-VQA framework, we conducted comparative assessments with existing methods on two benchmark datasets, namely VQA-RAD and SLAKE. Accuracy on open-ended questions served as the performance metric. Tri-VQA leverages a stable mutual reasoning relationship between visual input (V), question (Q), and answer (A) to determine the most reasonable answer. As depicted in Table~\ref{tab:med}, our approach outperforms all other methods in terms of open-ended performance, particularly on more challenging questions, across both datasets.

\subsection{Ablation Studies}

\begin{figure}[!t]
    \centering
    \includegraphics[width=\linewidth]{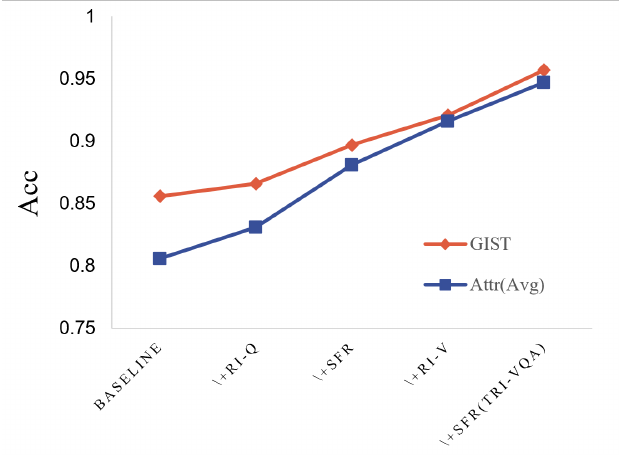}
    \caption{Tri-VQA Component Ablation Experiment.}
    \label{fig:ab}
\end{figure}

\subsubsection{Ablation Experiments of Tri-VQA Components}
Fig. \ref{fig:ab} illustrates the ablation experiment of Tri-VQA components, showing the significant contributions of Reverse Inference (RI) and Second Forward Reasoning (SFR) to the correctness of VQA, which cannot be ignored. The baseline represents the results obtained using a forward reasoning structure, which is the foundational framework adopted by the majority of existing Med-VQA approaches. The inclusion of reverse inference for question analysis has led to a significant improvement in the average inference accuracy for multiple attributes. During multi-attribute analysis, the need to address multiple attribute questions for the same image can result in confusion within the forward reasoning structure. Because linking different questions, the same image, and different answers poses a formidable challenge for VQA. However, reverse inference allows the model to backtrack and explore different questions that may correspond to different answers, providing a more robust linkage for VQA inference. We analyze this as the primary contributing factor to the performance enhancement. The incorporation of SFR not only ensures the semantic coherence of features obtained through reverse inference, thereby strengthening the network's ability for backward reasoning but also provides additional training for the forward reasoning pathway. With the incorporation of reverse inference on image features and the addition of SPR, a robust triangular reasoning structure, known as Tri-VQA, is established. This demonstrates that through increased reverse inference, the forward reasoning process, which infers the answers, can be strengthened in terms of rationality and correctness. This is because accurate reverse inference constrains the validity of forward reasoning.
\begin{figure}[!t]
    \centering
    \includegraphics[width=\linewidth]{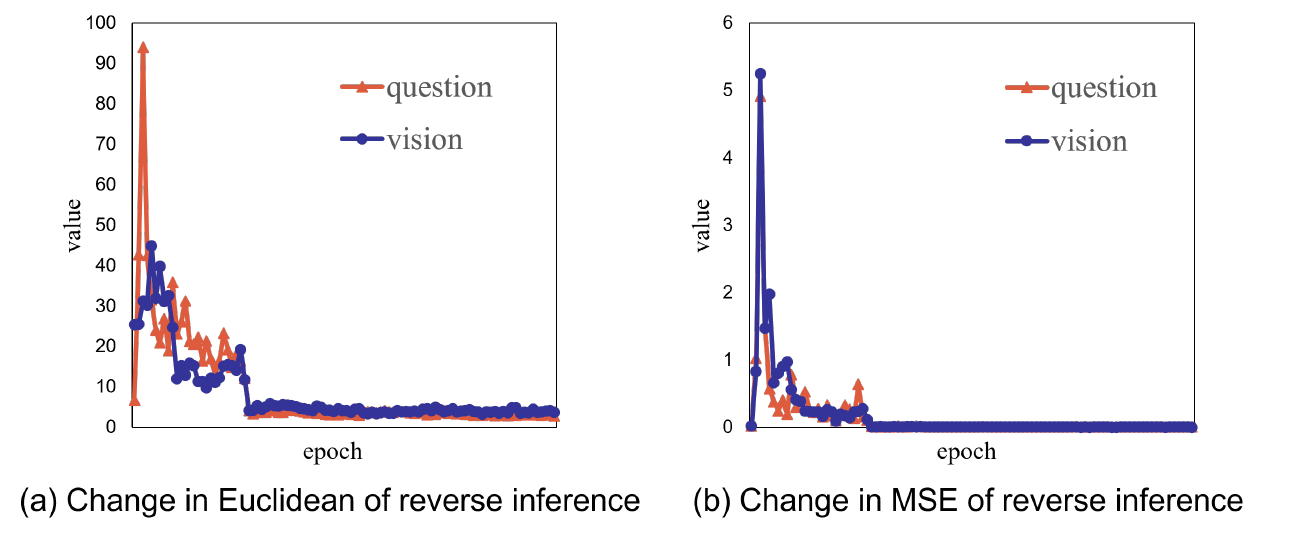}
    \caption{The change in similarity measurement metrics during training.}
    \label{fig:dis}
\end{figure}
\subsubsection{Validation of Inference Reliability}
Tri-VQA leverages reverse inference to enhance the reliability and coherence of generated answers. In this chapter, we validate the effectiveness of reverse inference in improving the reliability of answers produced by VQA by measuring the similarity between the features of generated images and questions during reverse inference and the features of real images and questions. 

Fig. \ref{fig:dis} illustrates the Mean Squared Error (MSE) values and Euclidean distances between the inferred features through reverse inference and the real features during the training process. The combined use of MSE and Euclidean distance allows for a comprehensive measurement of similarity between features from the perspectives of numerical differences and geometric distances, respectively. The results demonstrate that during the training process, the discrepancy between the inferred features and the ground truth features gradually diminishes. This trend is observed in both MSE and Euclidean distance metrics, highlighting the effectiveness of the inverse inference training.

We also analyzed the results of inverse inference for images and questions on the test data, and the specific results are presented in Table \ref{tab:dis}. The table displays the average measurements of similarity  between the inferred images and features and the ground truth features, categorized by different attributes. Based on the results obtained from the table, it can be observed that across all attributes, the results of reverse inference are consistently superior when the inference answer is correct compared to when it is incorrect. This provides strong evidence for the significance of reverse inference as a reliable means of assessing the correctness of answers. There are situations in which the disparity between reverse inference features is relatively small when predicting correct or incorrect answers. Intriguingly, in certain instances, the accuracy of reverse inference features is even higher when the answer prediction is incorrect compared to when it is correct. Take, for instance, the "Original" attribute, where reverse inference infers ${V}$. Interestingly, in cases of incorrect answer predictions, the MSE value is lower than that of correct answer predictions (MSE: 4.17 $<$ 4.21). We posit that this phenomenon can be attributed to the inherent difficulty in discerning the attribute of "Original" itself, as the images exhibit indistinguishable characteristic features, leading to the network's overconfidence. Similar challenges may be present to varying degrees in other attributes, where the network fails to discriminate even in cases of incorrect answer predictions using reverse inference. However, this observation underscores the value of reverse inference as an indicator of answer reliability in the majority of scenarios. The progressive improvement in GIST diagnosis results can be analyzed in light of the enhanced accuracy of multi-attribute recognition. The improvement in multi-attribute recognition contributes significantly to the accurate diagnosis of the final GIST. 

\section{Conclusion}
Our study proposes a novel triangle reasoning model, named Tri-VQA, based on tumor attributes. The model establishes a reverse causal reasoning relationship between the question (\textit{q}), image (\textit{v}), and answer (\textit{a}), enhancing the reasoning capability of Med-VQA. Additionally, by validating the correctness of reverse reasoning, it provides an evaluation metric for the reliability of the inferred answers. Extensive experiments are conducted on a multi-attribute EUS dataset and two publicly available Med-VQA datasets, demonstrating the excellent performance of Tri-VQA in reasoning. The analysis of reverse reasoning correctness provides a novel objective metric for assessing the reliability of answers generated by Med-VQA. This may bring new insights to Med-VQA and decision support in the field of medicine.

\section*{Acknowledgment}
This work is partially supported by the National Natural Science Foundation of China (62376231), the Sichuan Science and Technology Program (2023YFG0267), Natural Science Foundation of Sichuan Province (2023NSFSC1616), Fundamental Research Funds for the Central Universities (2682021ZTPY030, 2682022KJ045).






\bibliographystyle{IEEEtran}
\bibliography{TriVQA}

\end{document}